\documentclass[lettersize,journal]{IEEEtran}
\usepackage{amsmath,amsfonts}
\usepackage{array}
\usepackage[caption=false,font=normalsize,labelfont=sf,textfont=sf]{subfig}
\usepackage{stfloats}
\usepackage{url}
\usepackage{verbatim}
\usepackage{graphicx}
\usepackage{cite}
\usepackage{multirow}%
\usepackage{amsmath,amsfonts}%
\usepackage{amsthm}%
\usepackage{amssymb}
\usepackage{mathrsfs}%
\usepackage{xcolor}%
\usepackage{textcomp}%
\usepackage{manyfoot}%
\usepackage{booktabs}%
\usepackage{algorithm}%
\usepackage{algorithmicx}%
\usepackage{algpseudocode}%
\usepackage{listings}%

\usepackage{booktabs}
\usepackage{multirow}
\usepackage{CJKutf8}
\usepackage{xcolor}
\usepackage{hyperref}
\usepackage{array}
\usepackage{arydshln}
\usepackage{diagbox}
\usepackage{algorithm}
\usepackage{makecell}

\makeatletter
\def\adl@drawiv#1#2#3{%
        \hskip.5\tabcolsep
        \xleaders#3{#2.5\@tempdimb #1{1}#2.5\@tempdimb}%
                #2\z@ plus1fil minus1fil\relax
        \hskip.5\tabcolsep}
\newcommand{\cdashlinelr}[1]{%
  \noalign{\vskip\aboverulesep
           \global\let\@dashdrawstore\adl@draw
           \global\let\adl@draw\adl@drawiv}
  \cdashline{#1}
  \noalign{\global\let\adl@draw\@dashdrawstore
           \vskip\belowrulesep}}
\makeatother

\newcommand{\Mx}{\boldsymbol{\mathit{X}}} 
\newcommand{\My}{\boldsymbol{\mathit{Y}}} 
\newcommand{\Mw}{\boldsymbol{\mathit{W}}} 

\newcommand{\CalS}{\mathcal{S}}
\newcommand{\CalC}{\mathcal{C}}
\newcommand{\CalL}{\mathcal{L}}
\newcommand{\CalM}{\mathcal{M}}
\newcommand{\CalX}{\mathcal{X}}
\newcommand{\CalY}{\mathcal{Y}}
\newcommand{\CalA}{\mathcal{A}}

\newcommand{\SetC}{\mathbb{C}}
\newcommand{\SetD}{\mathbb{D}}
\newcommand{\SetX}{\mathbb{X}}

\newcommand{\SetR}{\mathbb{R}}
\newcommand{\SetY}{\mathbb{Y}}

\begin{document}

\title{Adapter-based Selective Knowledge Distillation for Federated Multi-domain Meeting Summarization}

\author{

Xiachong Feng, Xiaocheng Feng, Xiyuan Du, Min-Yen Kan, Bing Qin

\thanks{Corresponding author: Xiaocheng Feng}

\thanks{Xiachong Feng, Xiaocheng Feng, Xiyuan Du and Bing Qin are with the Research Center for Social Computing and Information Retrieval, Harbin Institute of Technology, 150001, Harbin, Heilongjiang, China (e-mail: {xiachongfeng,xcfeng,xydu,qinb}@ir.hit.edu.cn)}

\thanks{Xiaocheng Feng and Bing Qin are with the Peng Cheng Laboratory, 518000, Shenzhen, Guangdong, China (e-mail: {xcfeng,qinb}@ir.hit.edu.cn)}

\thanks{Min-Yen Kan is with the School of Computing, National University of Singapore, 117417, Singapore (e-mail: {knmnyn@nus.edu.sg})}

}

\maketitle

\begin{abstract}
Meeting summarization has emerged as a promising technique
for providing users with condensed summaries.
However, existing work has focused on training models on centralized data, neglecting real-world scenarios where meeting data are infeasible to collect centrally, due to their sensitive nature. 
This gap motivates us to explore federated learning for meeting summarization.
Two critical challenges impede progress.
First, state-of-the-art summarizers are based on parameter-heavy pre-trained models. Exchanging such a model's parameters across clients imposes large bandwidth costs.
Second, as real-world meeting data belong to various domains and are distributed across clients, they are instances of non-identically and independently distributed (non-IID). 
IID assumptions do not hold, which changes which forms of learning algorithms best apply.
To address this, we propose {\it Adapter-based Federated Selective Knowledge Distillation} (\textsc{AdaFedSelecKD}) for training performant client models. 
Specifically, we develop an adapter-based summarization model where two adapters cooperatively facilitate learning using fewer parameters to reduce communication costs.
Then, we devise a selective knowledge distillation strategy,
assisting clients in robustly handling domain-focused modelling on their own data, while leveraging global parameters based on non-IID data.
Extensive experiments on the QMSum benchmark demonstrate \textsc{AdaFedSelecKD} can achieve comparable performance with powerful centralized training methods, and shows its generalizability and robustness.
\end{abstract}

\begin{IEEEkeywords}
Meeting Summarization, Federated Learning, Knowledge Distillation, Parameter-efficient Fine-tuning.
\end{IEEEkeywords}

\section{Introduction}
\IEEEPARstart{M}{eeting} summarization aims to produce concise meeting summaries given lengthy meeting transcripts, efficiently facilitating readers to grasp essential meeting information \cite{Banerjee2015GeneratingAS}. 
With the advancement of meeting technologies, many meetings are now also recorded regularly and automatically transcribed with AI tools, facilitating offline meeting reviews.
Meeting summarization can leverage these inputs, further building capabilities to mitigate meeting overload. 

Meeting summarization has attracted extensive research attention as of late \cite{Kumar2022MeetingSA,Rennard2022AbstractiveMS}. Existing endeavours focus on developing performant summarization models, utilizing data resources located in a single location (also known as {\it centralized meeting summarization}). \cite{Feng2020DialogueDG,Zhu2020AHN,Zhong2021QMSumAN,Liu2022DynamicSW,Lee2023PrivateMS,Hu2023MeetingBankAB,wu2023vcsum}.

While meaningful in theory, in practice real-world meeting summarization has additional privacy challenges that substantially change the problem framing.  
Concretely speaking, real-world meetings inextricably contain highly private and sensitive information; e.g.,  confidential company contents and personal information that are private\cite{mccosker2001undertaking}. 
When extended to multi-modal data, video and audio meeting recordings often also meet with facial representation and voiceprint issues since both are likewise highly sensitive \cite{jain2000biometric}.
For these reasons, meeting data is highly sensitive and unable to be shared for model training purposes and is typically siloed. This makes the collection of meeting data in a central location infeasible.

As such, despite the encouraging research achievements reported in the current literature, we find such solutions do not meet the requirements of real-world scenarios. They neglect the investigation towards developing solutions where meeting data are necessarily siloed and are distributed across different client sites.

\begin{figure}[t]
	\centering
	\includegraphics[scale=0.54]{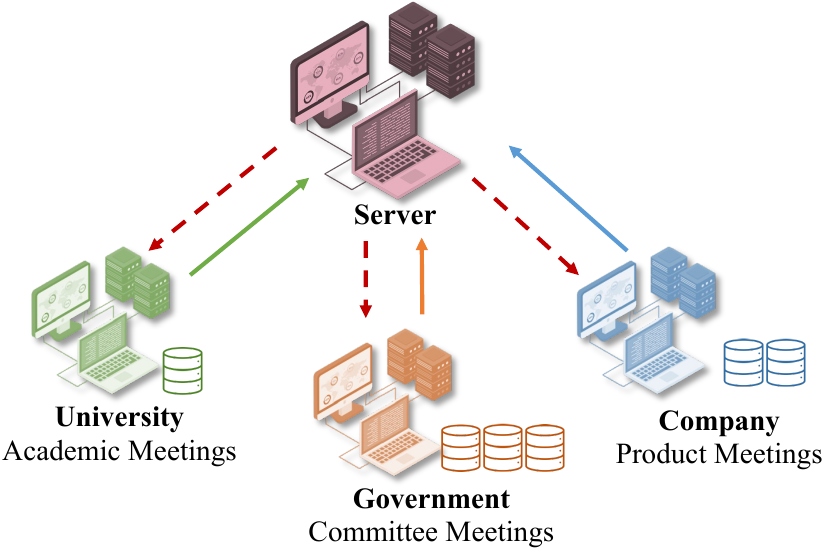}
	\caption{The overall federated learning framework of multi-domain meeting summarization. In the concrete setting for this paper, there is one central server and three clients covering distinct domains: \textit{Academic}, \textit{Committee} and \textit{Product}. Each client uniquely maintains its own domain-specific data.}
	\label{fig:intro}
\end{figure}

To close the above gap, we take the first step to study the meeting summarization task by leveraging a \textit{federated learning framework}, a widely-adopted approach for decentralized machine learning.  It enables model training across multiple distributed clients \cite{Yang2019FederatedML,Liu2021FederatedLM}. 
Figure~\ref{fig:intro} depicts the entire learning framework that aims to effectively train performant client-side summarization models by deriving global knowledge from other clients, without needing to access their private data. 
However, there are two critical challenges that need to be carefully addressed in order to learn high-performance summarization models under federated learning.
First, current state-of-the-art meeting summarization models are based on pre-trained language models that maintain a very large number of parameters.  Updating all model parameters represents an infeasible communication cost.
Instead, \textit{limited scale client--server communication} is more realistic. This restricts the exchange of parameter updates between the server and its clients to a budget.
Second, meetings distributed across multiple clients often belong to different domains. 
Figure~\ref{fig:intro} illustrate this scenario, in which there exists three meeting domains: academic, committee, and product, respectively. A single, central model would not serve to support the distinct needs of the different domains. 
This challenging \textit{non-identically and independently distributed (non-IID)} data learning setting often causes the client model to deviate from its own domain as it learns global knowledge based on non-IID data.

To mitigate the above two challenges, we propose a unified method, dubbed Adapter-based Federated Selective Knowledge Distillation (\textsc{AdaFedSelectKD}).
To address the first challenge, we draw support from parameter-efficient fine-tuning techniques and design an adapter-based summarization model to reduce communication costs.
Specifically, we introduce a few lightweight trainable adapters \cite{Houlsby2019ParameterEfficientTL} to pre-trained language models \cite{Lewis2019BARTDS,Beltagy2020LongformerTL} while keeping the pre-trained language models frozen.
We meticulously design two types of adapters --- global adapter and local adapter --- tailored for the federated learning framework to facilitate information exchange between the server and clients.
In particular, the global adapter is responsible for providing global knowledge while local adapters are optimized towards the local meeting summarization task. 
To address the second challenge, we devise a federated selective knowledge distillation strategy to not only effectively derive global knowledge for the client summarization model, but also train the model to favour its own local domain performance.
Specifically, the client model adopts knowledge distillation \cite{Gou2020KnowledgeDA} as the optimization algorithm to both learn from its local data and distill global knowledge from the global adapter. 
Moreover, we propose an entropy-based selective strategy based on the assumption that the higher the entropy of global knowledge, the more uncertain the knowledge. 
This adaptively distills knowledge from the global adapter.

We conduct experiments on the QMSum benchmark \cite{Zhong2021QMSumAN}, which comprises meeting summarization data across three distinct domains: academic, committee and product.
The automatic evaluation results based on three model variants across three clients consistently demonstrate the efficacy of our proposed method.  Our results achieve comparable results to centralized training methods. 
Moreover, human evaluation results validate the substantial improvements attained by our method over baseline approaches.
We further conduct downstream analyses of our model's various settings that allow us to conclude that our method is both generalizable and robust.

\begin{table}[t]
\caption{Mathematical notations utilized in this paper.}
\label{tab:natation}
\centering
    \begin{tabular}{cl}
        \toprule 
        \textbf{Notation} & \textbf{Description} \\
        \midrule
        \multicolumn{2}{l}{Dataset-related} \\
        \midrule
        $\SetD$ & meeting summarization dataset \\
        $\SetX$ & a set of input documents   \\
        $\SetY$ & a set of meeting summaries  \\
        $\CalX$ & one input document $\CalX \in \SetX$  \\
        $\CalY$ & one meeting summary $\CalY \in \SetY$   \\
        $V$ & vocabulary \\
        \midrule
        \multicolumn{2}{l}{Framework-related} \\
        \midrule
        $\CalS$ & server \\
        $\CalC$ & client \\
        $i$ & index of the client \\
        $\SetC$ & a set of clients, $\CalC_i \in \SetC$ \\
        ${\CalA}_c$ & client-side optimization algorithm \\
        ${\CalA}_s$ & server-side aggregation algorithm  \\
        \midrule
        \multicolumn{2}{l}{Model-related} \\
        \midrule
        ${\CalM}$ & client-side task model (meeting summarizer)  \\
        $\Mw$ & learnable parameters of the model ${\CalM}$  \\
        $\Mx$ & output representation of the encoder layer  \\
        $\My$ & output representation of the decoder layer  \\
        ${\hat{\My}}$ & output representation of the adapter  \\
        $l$ & index of the transformer layer \\
        $L$ & the number of transformer layers \\
        $n$ & the dimension of the model ${\CalM}$ \\
        $m$ & adapter bottleneck dimension  \\
        $q$ & normalized output distribution (after \texttt{softmax}) \\
        \midrule
        \multicolumn{2}{l}{Learning-related} \\
        \midrule
        $\CalL$ & loss function \\
        $\lambda$ & weight for the knowledge distillation loss $\CalL_{KD}$ \\
        $r$ & index of the federated learning round \\
        $H({q})$ & entropy of the distribution ${q}$ \\
        $\tau$ & threshold for the entropy \\
        \bottomrule
    \end{tabular}
\end{table}

\section{Preliminaries}
We first introduce the multi-domain meeting summarization dataset and the task definition, then provide an overview of federated learning.  We define all of the mathematical notation employed in this work in Table~\ref{tab:natation}.

\subsection{Multi-domain Meeting Summarization Dataset}

\begin{table*}[t]
\setlength\tabcolsep{5pt}
\caption{Statistics for the QMSum dataset encompass three domains: Academic, Committee and Product. ``\#" indicates the quantity of document--summary pairs. The average number of turns during meetings is denoted by ``Avg. Turns''.  ``Avg. Speakers'' represents the average number of speakers participating in the meetings. ``Avg. Tokens'' refers to the average number of words spoken during the meetings, and ``Avg. Sum'' indicates the average number of words in the summaries.}
\label{tab:datasets}
\centering
    \begin{tabular}{lrrrrrrrrr}
        \toprule 
        & \multicolumn{3}{c}{\textbf{Academic}} &  \multicolumn{3}{c}{\textbf{Committee}} & \multicolumn{3}{c}{\textbf{Product}} \\
        \cmidrule(r){2-4}\cmidrule(r){5-7}\cmidrule{8-10}
        & \textbf{Train} & \textbf{Valid} & \textbf{Test} & \textbf{Train} & \textbf{Valid} & \textbf{Test} & \textbf{Train} & \textbf{Valid} & \textbf{Test}  \\
        \midrule
        \# &218  &45  &49  &284 &67 &66  &593 &125 &129   \\
        Avg. Turns  &54.64   &59.33  &46.45   &9.51 &9.13 &10.85   &68.61 &79.30 &77.86  \\
        Avg. Speakers &4.35 &4.09 &4.22  &3.26 &3.10 &4.14  &3.76 &3.90 &3.93 \\
        Avg. Tokens &1049.8   &1156.53  &912.00  &667.14 &607.9 &713.26   &898.04 &903.53 &933.67 \\
        Avg. Sum  &46.22  &50.67  &43.82  &77.95 &73.67 &69.94 &65.09 &65.38 &57.38  \\
        \bottomrule
    \end{tabular}
\end{table*}

In this paper, we leverage the QMSum dataset \cite{Zhong2021QMSumAN} to conduct experiments under the federated learning setting.
QMSum consists of query--summary pairs over 232 meeting transcripts from three distinct domains: namely academic, committee and product meetings.  This dataset is thus well-suited for the multi-domain meeting summarization task.
Under our federated scenario, we posit that three clients hold meetings from each of the three distinct domains, respectively.
Notably, QMSum is a query-based meeting summarization dataset, in which each instance is composed of a specific query, the relevant meeting transcripts and the corresponding summary.
Following Lee and Sogaard \cite{Lee2023PrivateMS}, we concatenate the query and the meeting transcripts to construct the input document $\CalX$, resulting in a parallel corpus ${\SetD}:({\CalX},{\CalY}) \in ({\SetX},{\SetY})$, where $\CalY$ is the corresponding summary with respect to ${\CalX}$.
We give detailed statistics for the QMSum dataset in Table~\ref{tab:datasets}.

\subsection{Task Definition}
Given the document $\CalX$, the client-side meeting summarization model aims to produce a concise meeting summary $\CalY$, where $\CalX$ is the concatenation of one query's words and relevant meeting transcripts $[\underbrace{x_1, x_2,..., x_i}_{\texttt{query}}\texttt{\#SEP\#} \underbrace{x_{i+1}, x_{i+2}, ..., x_{|\CalX|}}_{\texttt{transcripts}}]$, \texttt{\#SEP\#} denotes one specific token between the query and transcripts. 
Note that speaker roles, such as ``marketing" and ``project manager", are treated as ordinary tokens and included in transcripts.
$\CalY$ consists of $|\CalY|$ words $[y_1, y_2, ..., y_{|\CalY|}]$. A brief example is shown as follows:
\begin{itemize}
    \item \textbf{Query $\CalX_{[1:i]}$}: Summarize the discussion about the trends of current remote controls. 
    \item \textbf{Meeting Transcripts $\CalX_{[i+1:|\CalX|]}$}: Marketing: This is just a presentation on the trends that we're gonna use to make the product stand out from ...... Project	Manager: What do you think of adding an LCD? ......
    \item \textbf{Corresponding Summary $\CalY$}: The group discussed different trends based on different ages of people, ..., finally they decide to add an LCD screen.
\end{itemize}
       
\subsection{Federated Learning Framework}
We investigate the multi-domain meeting summarization task under the federated learning framework.
Federated learning adopts a client--server paradigm and enables the collaborative training of models across multiple decentralized data sources without harvesting the sensitive raw data \cite{Yang2019FederatedML}.

Roughly speaking, the federated learning methodology progresses in a synchronous, iterative fashion.  
During each learning round, every client locally optimizes its own \textbf{\textit{client-side model}} based on its private data via the \textbf{\textit{client-side optimization algorithm}} and then transmits the updated parameters to the central server.
Subsequently, the server gathers the updates from clients and aggregates them into new server parameters by means of the \textbf{\textit{server-side aggregation algorithm}}.
Finally, the new global parameters are broadcast from the server to all clients for the next round.
It should be noted that three crucial components constitute this learning process:
\begin{itemize}
    \item \textit{\textbf{Client-side model}} ${\CalM}$ housing the client-specific model parameters $\Mw$.  It is in charge of generating meeting summaries.
    \item \textit{\textbf{Client-side optimization algorithm}} ${\CalA}_c$ endeavors to optimize the client-side model ${\CalM}$ based on the local private data ${\SetD}$.
    \item \textit{\textbf{Server-side aggregation algorithm}} ${\CalA}_s$ is responsible for aggregating parameters furnished by clients.
\end{itemize}

Formally speaking, there exists a set $\SetC$ of $|\SetC|$ clients, where each individual client is denoted $\CalC \in \SetC$. As illustrated in our concrete scenario in Figure~\ref{fig:intro}, there are three ($|\SetC|=3$) clients in our set; namely, the academic client, the committee  client, and the product client. Each client $\CalC_i$ possesses its own private domain-specific corpus of meeting summaries ${\SetD}_i$, as well as a client-specific meeting summarization model ${\CalM}_i$. The learnable parameters ${\Mw}_{i}$ of model ${\CalM}_i$ are optimized using the client-side optimization algorithm ${\CalA}_c$ based on the local dataset ${\SetD}_i$ in the $r_{th}$ round of optimization.
\begin{equation}
    {\Mw}_{i}^{r+1} \leftarrow {\CalA}_c({\CalM}_i({\Mw}_{i}^r),{\SetD}_i).
\end{equation}

Subsequently, the central server $\CalS$ aggregates all updated parameters ${\Mw}^{r+1}_i$ from the clients and adopts a server-side aggregation algorithm (${\CalA}_s$) to consolidate the information. In particular, we utilize the Federated Averaging (FedAvg) algorithm \cite{McMahan2016CommunicationEfficientLO} as our aggregation algorithm ${\CalA}_s$.
\begin{equation}
\begin{split}  
    {\Mw}^{r+1} \leftarrow  \sum_{i=1}^{|\SetC|} \frac{|{\SetD}_i|}{|{\SetD}|} {\Mw}_{i}^{r+1}, \; \text{where} \; |{\SetD}| = \sum_{i=1}^{|\SetC|} |{\SetD}_i|,
\end{split}
\end{equation}
where $|\SetC|$ denotes the number of clients, $|{\SetD}_i|$ represents the number of instances in the local dataset ${\SetD}_i$ and $|{\SetD}|$ gives the total number of instances among all clients.

Afterwards, the newly-gathered server-side parameters ${\Mw}^{r+1}$ are distributed to all clients $\CalC_i$ to offer enriched global knowledge.
In the forthcoming methodology section (\S\ref{sec:method}), we demonstrate our  contribution towards a more \textbf{communication-efficient client-side model ${\CalM}_i$} and \textbf{robust client-side optimization algorithm ${\CalA}_c$}.

\begin{figure*}[t]
    \centering
    \includegraphics[scale=0.55]{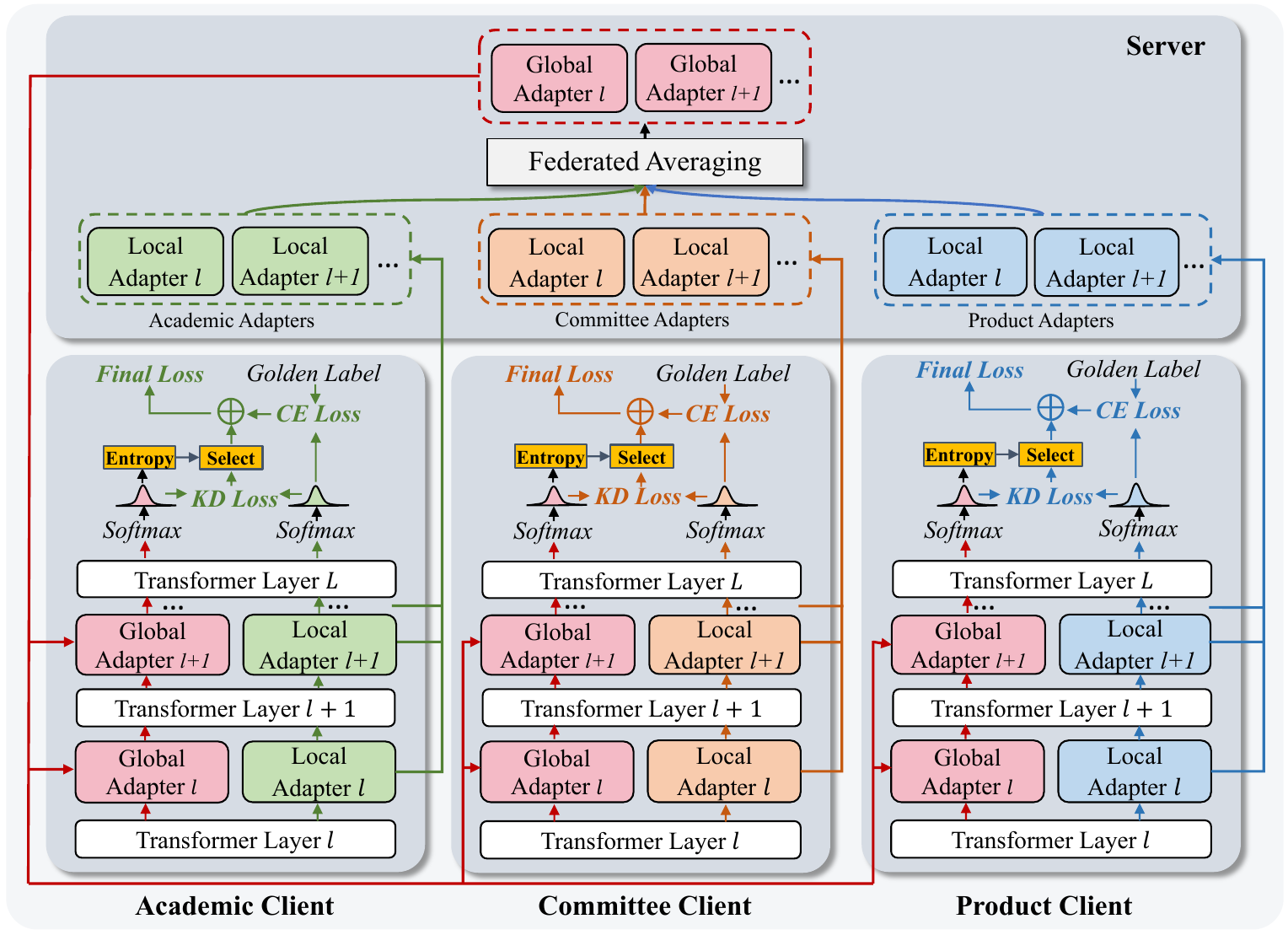}
    \caption{Illustration of our proposed \textsc{AdaFedSelectKD} learning framework. The overall framework adheres to a client--server learning paradigm.
    At the bottom, three clients are depicted, where each client adopts the selective knowledge distillation algorithm to optimize its own adapter-based meeting summarizer using its domain-specific private data. 
    Two types of adapters are tailored for the information exchange between the server and clients, including the global adapter and the local adapter.
    The optimized parameters from three clients are then conveyed to the central server.
    At the top, the central server employs the federated averaging algorithm to aggregate client information. The resulting new parameters are distributed to the clients for the subsequent learning round.
    }
    \label{fig:framework}
\end{figure*}

\section{Methodology}\label{sec:method}
We invent an integrated method, adapter-based federated selective knowledge distillation (\textsc{AdaFedSelectKD}), to achieve the efficient and robust federated multi-domain meeting summarization task. 
It comprises (1) A client-side model ${\CalM}$, which is an adapter-based meeting summarizer, and (2) A client-side optimization algorithm ${\CalA}_c$: which is a selective knowledge distillation algorithm. The overall \textsc{AdaFedSelectKD} learning procedure is illustrated in Figure \ref{fig:framework}.

\subsection{Overview}

Our proposed \textsc{AdaFedSelectKD} method significantly enhances two dimensions of the overall learning process.
\begin{itemize}
    \item \textit{\textbf{At the model level}}, we introduce a client-side model called the adapter-based meeting summarizer (${\CalM}$), which employs a frozen pre-trained language model as its backbone model (refer to \S\ref{sec:bone}) and integrates few learnable lightweight adapters to facilitate communication-efficient learning (refer to \S\ref{sec:ada}). Specifically, we develop two kinds of adapters tailored for the federated learning setting, namely the global adapter and the local adapter. The global adapter functions as an intermediary for exchanging information between the server and clients, while the local adapter not only distils global knowledge from the global adapter but also is optimized towards the local domain.
    \item \textit{\textbf{At the algorithm level}}, we devise a client-side optimization algorithm termed the selective knowledge distillation strategy (${\CalA}_c$), which adaptively and robustly optimizes local learnable adapters. Concretely, the knowledge distillation method permits deriving global knowledge from the server while ensuring the summarizer is prone to the local domain (refer to \S\ref{sec:kd}). Furthermore, our meticulously designed selective strategy draws support from entropy as a measure of uncertainty and shows great promise in transferring credible global knowledge to clients (refer to \S\ref{sec:ss}).  
\end{itemize}

\subsection{Adapter-based Meeting Summarizer}

In this section, we elucidate our motivation for incorporating adapters and subsequently delineate the precise model architecture for both the backbone model as well as two varieties of adapters, namely the global adapter and the local adapter. 

\subsubsection{Motivation}

In this section, we elaborate on the motivation underlying the design of our adapter-based meeting summarization system by addressing the following two questions:

\begin{itemize}
    \item\textbf{\textit{Why do we employ the adapter?}} In recent years, pre-trained language models have dominated the natural language processing field and have achieved remarkable success.
    Therefore, it is ideal to leverage such models as potent meeting summarization systems.
    However, there are two key challenges.
    Firstly, exchanging the parameters of these pre-trained language models incurs \textit{high client--server communication costs} due to a large number of model parameters. 
    Secondly, the \textit{lack of sufficient hardware capabilities} in real-world scenarios means some clients may struggle to handle such compute-intensive tasks.
    On this account, we apply the parameter-efficient fine-tuning strategy to the pre-trained language model by fine-tuning only a few lightweight adapters, thereby addressing the above challenges.
    \item \textbf{\textit{Why do we design two types of adapters?}} 
    The most primitive and widely adopted federated learning algorithm is the Federated Averaging, which directly broadcasts newly aggregated global parameters to clients to initialize their models for the next round of training \cite{McMahan2016CommunicationEfficientLO}.
    However, such a method notoriously performs poorly when clients hold non-IID data since aggregating divergent model parameters leads to model distraction \cite{Mora2022KnowledgeDF}, thereby leading to the client model cannot focus on its own domain.
    To combat this issue, we design two types of adapters.
    One is the \textit{global adapter} that receives server parameters and provides global knowledge via output distribution. The other is the \textit{local adapter} that is optimized towards the local summarization task. By bifurcating the parameters in this fashion, we overcome the difficulties that arise from aggregating disparate model parameters across clients with non-IID data.
\end{itemize}

\subsubsection{Backbone Model}\label{sec:bone}
We employ two types of pre-trained language models, one is BART \cite{Lewis2019BARTDS} and the other is LED \cite{Beltagy2020LongformerTL}, as the backbone model.
Both of them adopt the Transformer architecture \cite{Vaswani2017AttentionIA} and have been pre-trained on a huge volume of data.
They inherit a sequence-to-sequence framework, whereby the encoder first encodes the source sequence into distributed representations, which are then decoded by the decoder to generate the target summary.

Formally speaking, the input to the encoder is $\Mx^0$, which denotes the sum of the word embeddings $\Mx_{\mathrm{emb}}$ and position embeddings $\Mx_{\mathrm{pos}}$ of the input document $\CalX$. $\stackrel{L}{\underset{l=1}{:=}}$ symbolizes $L$ identical layers and $\Mx^{l-1}$ signifies the output representation of the ${l-1}_{th}$ encoder layer. Besides, $\textsc{Ffn}(\cdot)$ represents a position-wise feed-forward network, and $\textsc{Self-Att}(\cdot)$ denotes a multi-head self-attention.
\begin{equation}
\begin{split}
 \Mx^L &= \textsc{Encoder}(\Mx^0) \stackrel{L}{\underset{l=1}{:=}} \textsc{Ffn}\left(\textsc{Self-Att}(\Mx^{l-1})\right) 
\end{split}
\label{eq:trans_encoder}
\end{equation}

The decoder takes the output $\Mx^L$ of the encoder and the shifted right representation $\My^0$ of $\CalY$ as the input to produce the final representation $\My^L$, which will be projected into the vocabulary space in order to predict the summary.
\begin{equation}
\begin{split}
  \My^L &= \textsc{Decoder}(\My^0,\Mx^L)\\
  &\stackrel{L}{\underset{l=1}{:=}} \textsc{Ffn}\left(\textsc{Cross-Att}\left(\textsc{Self-Att}(\My^{l-1}), \Mx^L\right)\right)
\end{split}
\label{eq:trans_decoder}
\end{equation}

where $\textsc{Cross-Att}$ represents multi-head cross-attention. Additionally, each encoder and decoder layer is surrounded by residual connection \cite{He2015DeepRL} and layer normalization \cite{Ba2016LayerN}.

\subsubsection{Global-Local Adapters} \label{sec:ada}
Adapters are additional modules interpolated between layers of a pre-trained model\footnote{We conduct preliminary experiments and find that it is more effective to only add adapters between decoder layers of the pre-trained language model for the meeting summarization task. Similar conclusions are corroborated by Dai et al. \cite{Dai2022StableMoESR}.}.

Note that the core attribute of adapters is the exceedingly small number of parameters compared with the entire pre-trained language model, which paves the way for efficient fine-tuning and communication cost reduction. 
Specifically, we craft two types of adapters tailored for the federated learning framework.
\begin{itemize}
    \item \textit{\textbf{Global adapter}} plays the role of parameter container, which receives aggregated parameters from the server and generates the output distribution that provides global knowledge to the local client. Note that the global adapter is only responsible for passing parameters and will not be optimized.
    \item \textit{\textbf{Local adapter}} combined with the pre-trained language model servers as the final client meeting summarization model, which is core to mitigate the non-IID data learning challenge. Instead of directly adopting server parameters as local adapter parameters, the local adapter is optimized towards its local domain by training on  the local private dataset while deriving global knowledge from the global adapter.
\end{itemize}

\begin{figure}[t]
    \centering
    \includegraphics[scale=0.75]{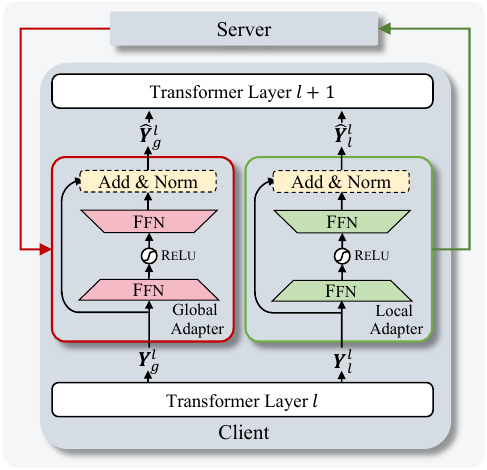}
    \caption{Illustration of the adapter architecture. Two types of adapters are added between transformer layers, including the global adapter and the local adapter. Both adapters share the same architecture, comprising a down-projection feed-forward layer, a non-linear activation function, an up-projection feed-forward layer and a residual connection module equipped with layer normalization.
    The global adapter receives parameters from the server and provides global knowledge, whereas the local adapter is co-optimized through training on the local data and distilling knowledge from the global adapter. The updated parameters are then transmitted to the server for the next round of learning.}
    \label{fig:adapter}
\end{figure}

Despite their distinct functions, the two types of adapters share an identical architecture. Specifically, we have adopted the bottleneck adapter architecture exemplified in  Houlsby et al. \cite{Houlsby2019ParameterEfficientTL} for our adapters. Precisely, each adapter consists of two feed-forward layers, one non-linear activation function, and a residual connection module with layer normalization. The overall architecture is illustrated in Figure \ref{fig:adapter}.

Formally speaking, given the adapter bottleneck dimension $m$, each adapter first utilizes a down-projection feed-forward module with learnable parameter ${\Mw}_{\text{down}} \in {\SetR}^{n \times m}$ to project the input ${\My}^{l} \in {\SetR}^{n}$ into the $m$-dimensional representation\footnote{We use the notation $\My$ since our adapters are added between transformer decoder layers.}, where ${\My}^{l}$ is produced by the $l_{th}$ transformer decoder layer of the pre-trained language model and $m$ is smaller than $n$.
Subsequently, a non-linear activation function $\texttt{ReLu}$ and an up-projection feed-forward module with learnable parameter ${\Mw}_{\text{up}} \in {\SetR}^{m \times n}$ are employed to project the vector back into $n$-dimensional representation.
Finally, a residual connection and layer normalization is applied to produce final ${\hat{\My}}^{l}$.
\begin{equation}
\begin{split}
    {\hat{\My}}^{l} \leftarrow \texttt{LayerNorm}({\My}^{l} + \texttt{ReLu}({{\My}^{l}}{{\Mw}_{\text{down}}}){\Mw}_{\text{up}})
\end{split}
\label{eq:adapter}
\end{equation}

Concretely, upon processing by the global adapter, ${\My}^{l}$ and ${\hat{\My}}^{l}$ are instantiated as ${\My}^{l}_{g}$ and ${\hat{\My}}^{l}_{g}$, respectively. 
Likewise, when processed by the local adapter, ${\My}^{l}$ and ${\hat{\My}}^{l}$ are instantiated as ${\My}^{l}_{l}$ and ${\hat{\My}}^{l}_{l}$, respectively.

\subsection{Selective Knowledge Distillation Strategy}

We first introduce our motivation to leverage knowledge distillation as the client-side optimization algorithm, and then present our selective strategy that further boosts the performance.

\subsubsection{Motivation}
Despite the apparent benefits of federated learning, the non-IID data learning setting leads to the domain drift problem of the client model.
In other words, directly using the global parameters derived by aggregating updates from distinct different domains makes it impossible for local models to focus on their own domain.
To remedy this issue, previous efforts have discovered knowledge distillation as one performant method \cite{Mora2022KnowledgeDF}. 
Building on this foundation, we first put forward our optimization method by dexterously unifying both the adapter-based parameter-efficient fine-tuning strategy and the knowledge distillation method, which not only reduces communication costs but also facilitates the robust learning of non-IID data across clients.
Additionally, global knowledge from the server is not always informative and beneficial \cite{He2022LearningCS}.
To address this concern, we devise a selective strategy, culminating in our final client-side optimization algorithm $\CalA_c$, \textsc{AdaFedSelectKD}, which adaptively and robustly distils credible global knowledge to the client model.
Algorithm \ref{alg:fed} shows the entire \textsc{AdaFedSelectKD} algorithm.

\begin{algorithm}[t]
\caption{Adapter-based Federated Selective Knowledge Distillation Algorithm. \\
$\SetC$ is the client set, $\Mw$ is the learnable parameters, $\SetD$ is the meeting summarization dataset, $\ensuremath{E}$ is the number of local epochs and $\eta$ is the learning rate.}\label{alg:fed}
\begin{algorithmic}[1]

\Procedure{Server executes:}{} 
    \State initialize ${\Mw}^1$
    \For{each round $r = 1, 2, \dots$}
     \For{each client $\CalC_i \in \SetC$ \textbf{in parallel}}
       \State ${\Mw}_{i}^{r+1} \leftarrow \textsc{AdaFedSelectKD}(i, {\Mw}^{r})$ 
     \EndFor
     \State $|{\SetD}| = \sum_{i=1}^{|\SetC|} |{\SetD}_i|$
     \State ${\Mw}^{r+1} \leftarrow \sum_{i=1}^{|\SetC|} \frac{|{\SetD}_i|}{|{\SetD}|} {\Mw}_{i}^{r+1}$ 
   \EndFor
\EndProcedure

\\

\Procedure{AdaFedSelectKD:}{$i, {\Mw}$}       
  \For{each local epoch from $1$ to $\ensuremath{E}$}
    \For{each training instance $(\CalX,\CalY) \in \SetD_i$}
        \For{each training target word $y_t \in \CalY$}
            \State $q_g, q_l = {\CalM_i}(\Mw, (\CalX,\CalY_{[1:y_{t-1}]}))$ 
            \If{$H({q_g}) < \tau$} 
                \State $\CalL=(1-\lambda)\CalL_{CE}\left(q_l, y_t\right)+\lambda\CalL_{KL}\left(q_l, q_g\right)$
            \Else
                \State $\CalL = \CalL_{CE}\left(q_l, y_t\right)$
            \EndIf
            \State $\Mw \leftarrow \Mw - \eta \triangledown {\CalL}(\Mw; (\CalX,\CalY_{[1:y_{t-1}]}))$
        \EndFor
    \EndFor
 \EndFor
 \State return ${\Mw}$ to server
\EndProcedure
\end{algorithmic}
\end{algorithm}

\subsubsection{Knowledge Distillation}\label{sec:kd}
Knowledge distillation is a solid method for transferring knowledge of the teacher model to the student model by minimizing the discrepancy between the outputs from two models with a proxy dataset \cite{Hinton2015DistillingTK}.

Formally speaking, given the training set ${\SetD}$, for each training instance $({\CalX},{\CalY}) \in ({\SetX},{\SetY})$, we obtain the final-layer representations, ${\hat{\My}}^{L}_{g}$ and ${\hat{\My}}^{L}_{l}$, produced by the global adapter and the local adapter, respectively.
After being transformed by the language head, which projects the representation into $|V|$-dimensional probability distributions (after \texttt{softmax} operation), we obtain the outputs $q_g$ and $q_l$, respectively.
Within our framework, we regard $q_g$ as the output of the teacher model and $q_l$ as the output of the student model.
Consequently, the local adapter can be trained utilizing a linear combination of two loss functions.
\begin{equation}
\begin{split}
\CalL=(1-\lambda) \CalL_{CE}\left(q_l, y\right)+\lambda \CalL_{KL}\left(q_l, q_g\right)
\end{split}
\label{eq:kd_loss}
\end{equation}
where $\CalL_{CE}$ represents the cross-entropy loss between the predicted distribution $q_l$ and the one-hot true label $y$. $\CalL_{KL}$ denotes the Kullback-Leibler divergence between $q_g$ and $q_l$. The scalar $\lambda$ serves to determine the weight between the two loss terms in the overall objective function.

\subsubsection{Selective Strategy}\label{sec:ss}
The immaturity parameters provided by the server inevitably introduce useless information to local model learning \cite{Qi2023BetterGR}.
To alleviate this problem, we draw inspiration from previous works in classification \cite{gal2016dropout} and summarization \cite{Poel2022MutualIA}, which employ the \textit{entropy} as a measure of uncertainty, and devise a selective strategy to adaptively distill the knowledge provided by the server.

In detail, when training the $t_{th}$ target word $y_t$ of the instance $(\CalX,\CalY)$, we have a normalized $|V|$-dimensional probability distribution
${q_g} = [{q_g^1}, {q_g^2},...,{q_g^{|V|}}]$, where $|V|$ is the vocabulary size. Given this, the entropy of ${q_g}$ is defined as:
\begin{equation}
\begin{split}
H({q_g})=-\sum_{i=1}^{|V|} P\left({q_g^i}\right) \log P\left({q_g^i}\right)
\end{split}
\label{eq:entropy}
\end{equation}

We assume that \textbf{\textit{the higher the entropy, the more uncertain the knowledge provided by the global adapter}}, which means global knowledge with high entropy has no confidence in handling the current learning situation, thereby needing to be ignored.
Based on this assumption, we finally propose our selective knowledge distillation strategy whereby the knowledge distillation loss is only accounted for when the entropy falls below a pre-defined entropy threshold $\tau$. 
\begin{equation}
\begin{split}
\CalL = \begin{cases}(1-\lambda) \CalL_{CE}\left(q_l, y_t\right)+\lambda \CalL_{KL}\left(q_l, q_g\right) & \text { if } H({q_g}) < \tau  \\  \CalL_{CE}\left(q_l, y_t\right) & \text { otherwise. }\end{cases}
\end{split}
\label{eq:final_loss}
\end{equation}

\section{Experiments}
In this section, we first introduce our research questions and then present baseline methods including both non-federated and federated learning settings, and finally describe evaluation metrics and implementation details.

\subsection{Research Questions}
Our experiments are intended to address the following research questions:

\begin{itemize}
    \item \textbf{Research Question 1}: How does the proposed \textsc{AdaFedSelecKD} perform, and is it comparable to powerful centralized training methods? 
    \item \textbf{Research Question 2}: How well does the proposed \textsc{AdaFedSelecKD} generalize? Can it achieve good performance under a variety of settings, particularly under more severe non-IID data situations?
    \item \textbf{Research Question 3}: How does the proposed selective knowledge distillation strategy work specifically and what are the underlying mechanisms?
\end{itemize}

\begin{table*}[t]
\caption{Main results of adopting BART-large as the backbone model. The \textsc{AdaCentralized} is one super strong method that explicitly trains the model on data from all three domains. Client models obtained via different methods are tested on the domain-specific test set of each client. $\dagger$ and $\dagger\dagger$ indicate the first-ranked and second-ranked results respectively. Results are averaged over three random runs.}
\label{tab:main_res_bart_large}
\centering
    \begin{tabular}{cllccc}
        \toprule 
        \textbf{Client} & \textbf{Setting} & \textbf{Method} & \textbf{ROUGE-1} & \textbf{ROUGE-2} & \textbf{ROUGE-L}  \\
        \midrule
        \multirow{5}{*}{{{Academic}}} & Single & \textsc{AdaSingle} & 24.83  & 5.70  & 17.24  \\
        & Centralized & \textsc{AdaCentralized}  & \textbf{26.74}$^{\dagger\dagger}$  & \textbf{7.95}$^{\dagger}$ & \textbf{20.73}$^{\dagger}$ \\
        \cdashline{2-6}
        & Federated & \textsc{AdaFedAvg} & 25.76 & 6.18 & 18.70 \\
        & Federated & \textsc{AdaFedKD} & 26.66  & 7.02 & 19.26 \\
        & Federated & \textsc{AdaFedSelectKD}  & \textbf{27.09}$^{\dagger}$  &  \textbf{7.62}$^{\dagger\dagger}$ & \textbf{19.79}$^{\dagger\dagger}$  \\
        \midrule
        \multirow{5}{*}{{{Committee}}} & Single &\textsc{AdaSingle} & 32.38 & 13.08 & 23.24   \\
        & Centralized & \textsc{AdaCentralized}  & \textbf{34.59}$^{\dagger\dagger}$  & \textbf{15.79}$^{\dagger}$ & \textbf{25.32}$^{\dagger\dagger}$ \\
        \cdashline{2-6}
        & Federated & \textsc{AdaFedAvg} & 33.80 & 14.04 & 24.19 \\
        & Federated & \textsc{AdaFedKD} & 34.12 & 14.65 & 24.83 \\
        & Federated & \textsc{AdaFedSelectKD}  &  \textbf{34.70}$^{\dagger}$ & \textbf{15.19}$^{\dagger\dagger}$ & \textbf{25.37}$^{\dagger}$ \\ 
        \midrule
        \multirow{5}{*}{{{Product}}} & Single &\textsc{AdaSingle} & 31.83 & 11.11 & 21.27  \\
        & Centralized & \textsc{AdaCentralized}  & \textbf{34.53}$^{\dagger}$  & \textbf{12.78}$^{\dagger\dagger}$ & \textbf{23.71}$^{\dagger}$ \\
        \cdashline{2-6}
        & Federated & \textsc{AdaFedAvg} & 32.50 & 12.19 & 22.45 \\
        & Federated & \textsc{AdaFedKD} & 33.09  &  12.50 & 23.11 \\
        & Federated& \textsc{AdaFedSelectKD}  & \textbf{33.32}$^{\dagger\dagger}$ & \textbf{12.86}$^{\dagger}$ & \textbf{23.53}$^{\dagger\dagger}$ \\
        \bottomrule
    \end{tabular}
\end{table*}

\subsection{Baseline Methods}
Our baseline methods can be divided into two categories: non-federated learning and federated learning.  
All adopt the adapter-based pre-trained language model as the backbone model.

\begin{itemize}
    \item \underline{\textsc{AdaSingle}}.
        \\ \textbf{Setting}: Non-federated learning setting.
        \\ \textbf{Model}: The model only has the local adapter that will be optimized.
        \\ \textbf{Method}: Training and testing the model with data from a single domain.
    \item \underline{\textsc{AdaCentralized}}.
        \\ \textbf{Setting}: Non-federated learning setting.
        \\ \textbf{Model}: The model only has the local adapter that will be optimized.
        \\ \textbf{Method}: Training the model using the whole QMSum that covers all three domains and testing the model using only one single-domain data. \textit{Centralized training methods are always viewed as one super strong baseline for federated methods.} 
    \item \underline{\textsc{AdaFedAvg}}.
        \\ \textbf{Setting}: Federated learning setting.
        \\ \textbf{Model}: The model has one type of adapter that will be optimized during training.
        \\ \textbf{Method}: The fundamental federated learning algorithm, where clients hold client-specific parameters and perform local updates based on their private data via maximum likelihood estimation. Afterwards, the server gathers the weighted average of all client updates as the new global parameters, which will be distributed to all clients as new client-specific parameters.
    \item \underline{\textsc{AdaFedKD}}.
        \\ \textbf{Setting}: Federated learning setting.
        \\ \textbf{Model}: The model has two types of adapters including both the global adapter and the local adapter. The local adapter is optimized during training.
        \\ \textbf{Method}: Each client performs local updates via knowledge distillation while the server employs federated averaging as the parameter-aggregation algorithm.
    \item \underline{\textsc{AdaFedSelectKD}}.
        \\ \textbf{Setting}: Federated learning setting.
        \\ \textbf{Model}: The model has two types of adapters including both the global adapter and the local adapter. The local adapter is optimized during training.
        \\ \textbf{Method}: Based on the \textsc{AdaFedKD}, the selective strategy is introduced to filter out global knowledge.
\end{itemize}

\begin{table*}[t]
\caption{Main results of adopting BART-base as the backbone model. The \textsc{AdaCentralized} is one super strong method that explicitly trains the model on data from all three domains. Client models obtained via different methods are tested on the domain-specific test set of each client. $\dagger$ and $\dagger\dagger$ indicate the first-ranked and second-ranked results respectively. Results are averaged over three random runs.}
\label{tab:main_res_bart_base}
\centering
    \begin{tabular}{cllccc}
        \toprule 
        \textbf{Client} & \textbf{Setting} & \textbf{Method} & \textbf{ROUGE-1} & \textbf{ROUGE-2} & \textbf{ROUGE-L}  \\
        \midrule
        \multirow{5}{*}{{{Academic}}} & Single &  \textsc{AdaSingle} & 23.97  & 5.70  & 17.81  \\
        & Centralized & \textsc{AdaCentralized}  & \textbf{25.66}$^{\dagger}$  & \textbf{6.72}$^{\dagger}$ & \textbf{18.88}$^{\dagger\dagger}$ \\
        \cdashline{2-6}
        & Federated & \textsc{AdaFedAvg} & 24.19 & 6.19 & 18.20 \\
        & Federated &\textsc{AdaFedKD} & 24.84  & 6.25 & 18.32 \\
        & Federated &\textsc{AdaFedSelectKD}  & \textbf{25.25}$^{\dagger\dagger}$  & \textbf{6.48}$^{\dagger\dagger}$  & \textbf{18.98}$^{\dagger}$  \\
        \midrule
        \multirow{5}{*}{{{Committee}}} & Single & \textsc{AdaSingle} & 27.68  & 9.41  & 19.30  \\
        & Centralized & \textsc{AdaCentralized}  & \textbf{28.61}$^{\dagger\dagger}$  & \textbf{11.23}$^{\dagger}$ & \textbf{21.21}$^{\dagger}$ \\
        \cdashline{2-6}
        & Federated &\textsc{AdaFedAvg} & 28.09 & 9.87 & 20.31 \\
        & Federated &\textsc{AdaFedKD} &  28.21 & 10.40 & 20.49 \\
        & Federated &\textsc{AdaFedSelectKD}  & \textbf{28.88}$^{\dagger}$  & \textbf{11.14}$^{\dagger\dagger}$  &  \textbf{21.07}$^{\dagger\dagger}$ \\
        \midrule
        \multirow{5}{*}{{{Product}}} & Single &  \textsc{AdaSingle} & 28.47  & 9.86  & 19.85  \\
        & Centralized & \textsc{AdaCentralized}  & \textbf{30.83}$^{\dagger}$  & \textbf{11.15}$^{\dagger\dagger}$ & \textbf{20.92}$^{\dagger\dagger}$ \\
        \cdashline{2-6}
        & Federated & \textsc{AdaFedAvg} & 29.34 & 10.47 & 20.42 \\
        & Federated & \textsc{AdaFedKD} & 29.89  & 10.56 & 20.74 \\
        & Federated & \textsc{AdaFedSelectKD}  & \textbf{30.41}$^{\dagger\dagger}$ & \textbf{11.25}$^{\dagger}$ & \textbf{21.15}$^{\dagger}$ \\
        \bottomrule
    \end{tabular}
\end{table*}

\subsection{Evaluation Metrics}
We adopt the standard metrics ROUGE \cite{Lin2004ROUGEAP} for evaluation and obtain the $F_1$ scores for ROUGE-1, ROUGE-2, and ROUGE-L that measures the word-overlap, bigram-overlap and longest common sequence between the ground-truth and the generated summary respectively.
We use the implementation provided by HuggingFace\footnote{\url{https://github.com/huggingface/evaluate}}.

\subsection{Implementation Details}
We use the Flower framework to simulate the federated learning environment\footnote{\url{https://github.com/adap/flower}}.
Specifically, we establish one central server and three distributed clients for the academic, committee, and product domains, respectively. 
During each round of federated learning, all three clients are engaged in the training process, indicating a client participation rate of 100\%.
The server employs the federated averaging algorithm to aggregate the gathered information. 
On the client side, we employ both BART \cite{Lewis2019BARTDS} and LED \cite{Beltagy2020LongformerTL} as the backbone model to conduct experiments.
We initially conduct hyperparameter search experiments via grid search to determine the final hyperparameters.
For the BART-large model and LED-large model, adapters are added to the top six transformer decoder layers with an adapter bottleneck dimension of 2048.
For the BART-base model, adapters are added to the top three transformer decoder layers with an adapter bottleneck dimension of 1536.
For each client, we used the AdamW optimizer with a learning rate of 2e-4 and a batch size of 16. The weight decay is set to 0.01.
The loss weight $\lambda$ is set to 0.2 and the entropy threshold $\tau$ for the selective strategy is set to 5 across all model variants\footnote{Our codes and models will be made public.}.

\begin{table*}[t]
\caption{Main results of adopting LED-large as the backbone model. The \textsc{AdaCentralized} is one super strong method that explicitly trains the model on data from all three domains. Client models obtained via different methods are tested on the domain-specific test set of each client. $\dagger$ and $\dagger\dagger$ indicate the first-ranked and second-ranked results respectively. Results are averaged over three random runs.}
\label{tab:main_res_led_large}
\centering
    \begin{tabular}{cllccc}
        \toprule 
        \textbf{Client} & \textbf{Setting} & \textbf{Method} & \textbf{ROUGE-1} & \textbf{ROUGE-2} & \textbf{ROUGE-L}  \\
        \midrule
        \multirow{5}{*}{{{Academic}}} & Single & \textsc{AdaSingle} & 24.28  & 6.03  & 17.58  \\
        & Centralized & \textsc{AdaCentralized}  &  \textbf{26.30}$^{\dagger}$ & \textbf{6.94}$^{\dagger\dagger}$ & \textbf{18.64}$^{\dagger\dagger}$ \\
        \cdashline{2-6}
        & Federated & \textsc{AdaFedAvg} & 25.13 & 6.34 & 18.19 \\
        & Federated & \textsc{AdaFedKD} & 25.49  & 6.58  & 18.30 \\
        & Federated & \textsc{AdaFedSelectKD}  & \textbf{25.92}$^{\dagger\dagger}$  & \textbf{7.09}$^{\dagger}$  & \textbf{18.67}$^{\dagger}$  \\
        \midrule
        \multirow{5}{*}{{{Committee}}} & Single & \textsc{AdaSingle} & 32.69  & 13.03 & 22.58  \\
        & Centralized &\textsc{AdaCentralized} & \textbf{33.71}$^{\dagger}$ & \textbf{14.94}$^{\dagger\dagger}$  & \textbf{23.87}$^{\dagger}$  \\
        \cdashline{2-6}
        & Federated &\textsc{AdaFedAvg} & 32.72  & 13.88 & 23.22 \\
        & Federated &\textsc{AdaFedKD}  & 32.98  & 14.39  & 23.46 \\
        & Federated &\textsc{AdaFedSelectKD}  &  \textbf{33.53}$^{\dagger\dagger}$ & \textbf{14.95}$^{\dagger}$  & \textbf{23.85}$^{\dagger\dagger}$  \\
        \midrule
        \multirow{5}{*}{{{Product}}} & Single & \textsc{AdaSingle} & 30.12  &  9.85 & 19.70  \\
        & Centralized &\textsc{AdaCentralized}  & \textbf{32.72}$^{\dagger}$  & \textbf{10.75}$^{\dagger}$ & \textbf{22.17}$^{\dagger\dagger}$ \\
        \cdashline{2-6}
        & Federated &\textsc{AdaFedAvg} & 30.43 & 9.93  & 21.66 \\
        & Federated &\textsc{AdaFedKD} &  31.69 & 10.12 & 21.93 \\
        & Federated &\textsc{AdaFedSelectKD}  & \textbf{32.07}$^{\dagger\dagger}$  & \textbf{10.67}$^{\dagger\dagger}$  &  \textbf{22.64}$^{\dagger}$ \\
        \bottomrule
    \end{tabular}
\end{table*}

\section{Results}

\subsection{Research Question 1}
To answer the first research question ``\textit{How does the proposed \textsc{AdaFedSelecKD} perform, and is it comparable to powerful centralized training methods?}'', we conduct both automatic evaluations by comparing various methods and human evaluations to comprehensively access the performance.

\subsubsection{Automatic Evaluation}
The results illustrated in Tables \ref{tab:main_res_bart_large}, \ref{tab:main_res_bart_base} and \ref{tab:main_res_led_large} correspond to the BART-large, BART-based and LED-large backbone models, respectively.
To sum up, the following conclusions can be drawn.
Firstly, the outcomes confirm that our proposed \textsc{AdaFedSelectKD} outperforms the baseline method \textsc{AdaFedAvg}, improving the ROUGE score by approximately 1.2 points. 
Secondly, compared with \textsc{AdaFedKD}, our optimized \textsc{AdaFedSelectKD} demonstrates superior performance, which confirms that the selective strategy constitutes the vital component for robust and efficacious federated knowledge distillation.
Thirdly, the results also validate that our \textsc{AdaFedSelectKD} can achieve comparable even superior performance relative to \textsc{AdaCentralized}, which is a piece of solid evidence to verify the effectiveness of our method.
Fourthly, the improvements achieved across the three model variants indicate the stability and generalizability of our method.
\textbf{Due to the better results based on the BART-large backbone model, the following experiments are all based on the BART-large model}.

\begin{figure}[t]
    \centering
    \includegraphics[scale=0.23]{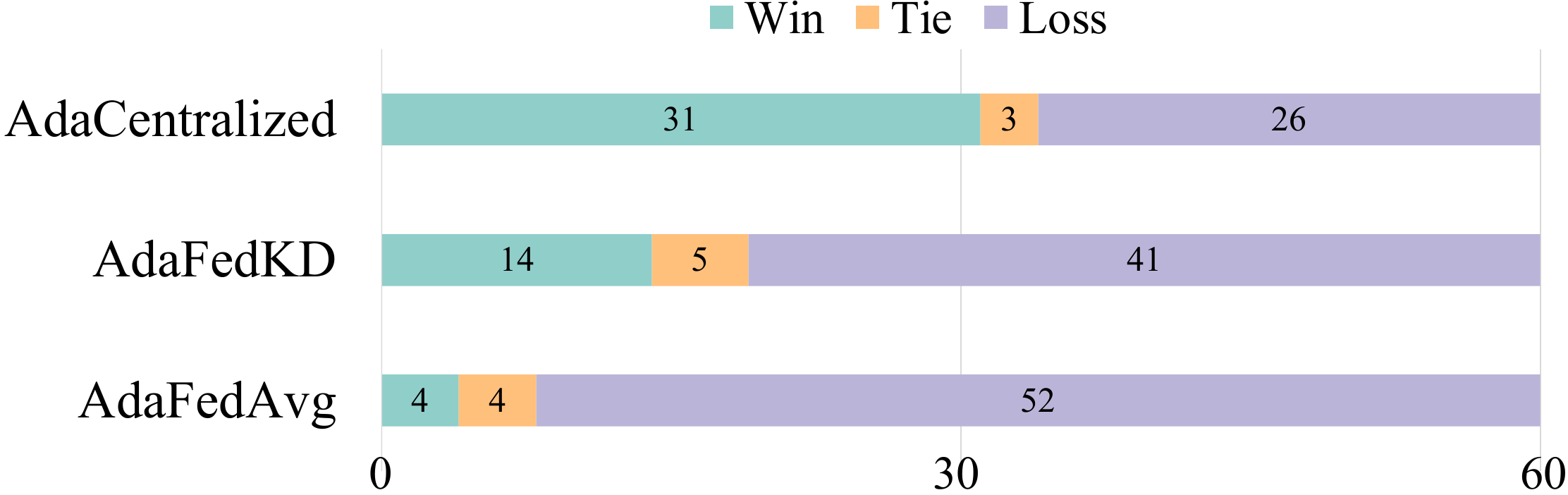}
    \caption{Generated meeting summary comparison of \textsc{AdaFedSelectKD} with other methods on 60 randomly-chosen meetings. For example, compared with \textsc{AdaFedSelectKD}, \textsc{AdaFedAvg} performs better on 4 of the 60 summaries and worse on 52.
 }
    \label{fig:human}
\end{figure}

\subsubsection{Human Evaluation}
We employ three evaluators to undertake our human evaluation.
All three evaluators are researchers in natural language processing who are well-versed in the task of meeting summarization.
Each evaluator is remunerated \$30 for this evaluation task.

First, 20 meetings are randomly selected per domain, amounting to 60 meetings in total.
For each meeting, we generate its meeting summary based on four methods: \textsc{AdaCentralized}, \textsc{AdaFedAvg}, \textsc{AdaFedKD} and \textsc{AdaFedSelectKD}. 
Each evaluator is provided with the meeting and a pair of summaries generated based on \textsc{AdaFedSelectKD} and another method respectively, in random order.
Ehe evaluator determines which summary is better (wins) or decides a tie between the two summaries according to domain expertise, informativeness and factual correctness of summaries.

We count the number of wins, ties and losses for each method, with the average results across the three evaluators (Figure~\ref{fig:human}). 
These observations point to the conclusion that our method exhibits an impressive win rate of up to 87\% vis-à-vis the baseline \textsc{AdaFedAvg} method.
It is noteworthy that \textsc{AdaFedSelectKD} achieves competitive results compared with the strong \textsc{AdaCentralized}, with a 43\%  win rate.
Additionally, the comparison with \textsc{AdaFedKD} also proves the necessity of our designed selective strategy.

\subsection{Research Question 2}
To answer the second research question ``\textit{How well does the proposed \textsc{AdaFedSelecKD} generalize? Can it achieve good performance under a variety of settings, particularly under more severe non-IID data situations?}'', we set up various experimental settings to provide more comparisons.

\subsubsection{IID and Balanced Data Setting}

\begin{figure}[t]
    \centering
    \includegraphics[scale=0.41]{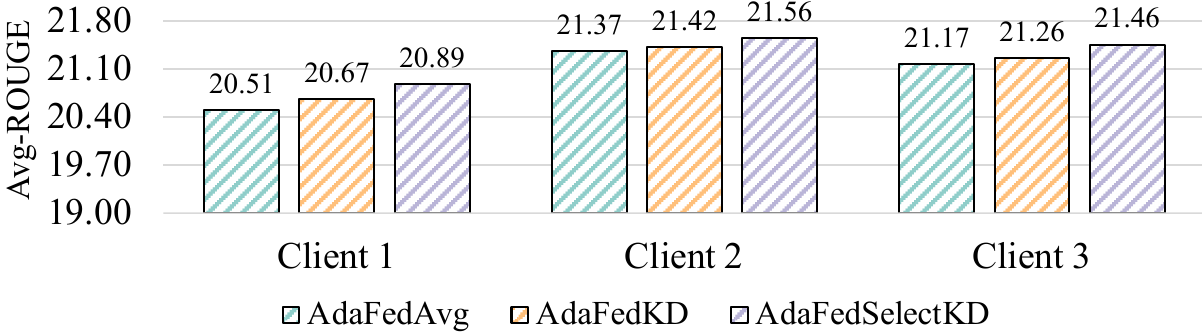}
    \caption{Average ROUGE results based on the IID and balanced data setting, where each client maintains meeting summarization data of the same distribution (IID) and holds the same amount of data instances (balanced).}
    \label{fig:iid}
\end{figure}

Under this setting, each client maintains meeting summarization data of the same distribution (IID), with equal quantities distributed across the three clients (balanced). 
First, we evenly divide the data of the three domains into three parts respectively.
Then, for the data of each domain, we distribute the three divided parts into the three clients respectively, resulting in our IID and balanced data setting, wherein each client holds one-third of the data in each of the three domains.
Subsequently, we conduct experiments leveraging \textsc{AdaFedAvg}, \textsc{AdaFedKD} and \textsc{AdaFedSelectKD} based on this newly curated data.
The results (shown in Figure \ref{fig:iid}) are averaged over three random runs, with the data being randomly re-divided for each run.
We find that given the IID and balanced data, all three clients demonstrate similar performance, with our \textsc{AdaFedSelectKD} being more effective compared with other federated baselines.
Despite the effectiveness of both \textsc{AdaFedKD} and  \textsc{AdaFedSelectKD}, we find they contribute marginally under this setting.
Our evidence reinforces the findings of previous works that federated knowledge distillation methods excel at overcoming the challenge of non-IID data learning but contribute little under the IID data setting.

\subsubsection{Non-IID and Balanced Data Setting}

\begin{figure}[t]
    \centering
    \includegraphics[scale=0.41]{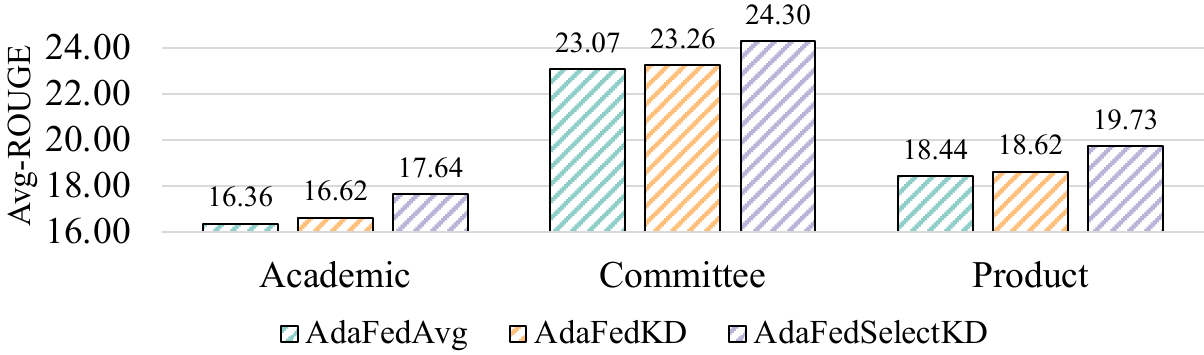}
    \caption{Average ROUGE results based on the non-IID and balanced data setting, where each client maintains domain-specific meeting summarization data (non-IID) and holds the same amount of data instances (balanced).}
    \label{fig:balanced}
\end{figure}

Under this setting, each client maintains domain-specific meeting summarization data (non-IID), with equal quantities distributed across the three clients (balanced). 
Specifically, for each domain, we randomly select 200 training instances, 40 validation instances and 40 testing instances from the corresponding QMSum portion, resulting in balanced data quantities across the three clients.
Subsequently, we conduct experiments using \textsc{AdaFedAvg}, \textsc{AdaFedKD} and \textsc{AdaFedSelectKD} based on this newly curated balanced data.
The results (depicted in Figure \ref{fig:balanced}) are averaged over three random runs, with data randomly re-selected for each run.
Firstly, it is evident that the ROUGE results show varying degrees of decline due to the reduction of data quantity relative to the full amount of data.
Secondly, under this well-formed setting, the utility of our \textsc{AdaFedSelectKD} is more fully exploited, with over 1 point ROUGE improvement directly compared with \textsc{AdaFedKD}.
We attribute this to the fact that the balanced data setting, which facilitates a consistent parameter optimization process on the client side, thus generating stable global parameters that allow our entropy-based selective strategy to make reliable distillation decisions. 
Thirdly, the results reveal that despite having the same amount of data across all three clients, the committee client attains superior performance. 
As indicated in Table~\ref{tab:datasets}, this can be attributed to the committee meeting's fewer turns and reduced input tokens, making it easier to train an effective summarizer.

\subsubsection{Extreme Non-IID and Unbalanced Data Setting}

\begin{figure}[t]
    \centering
    \includegraphics[scale=0.31]{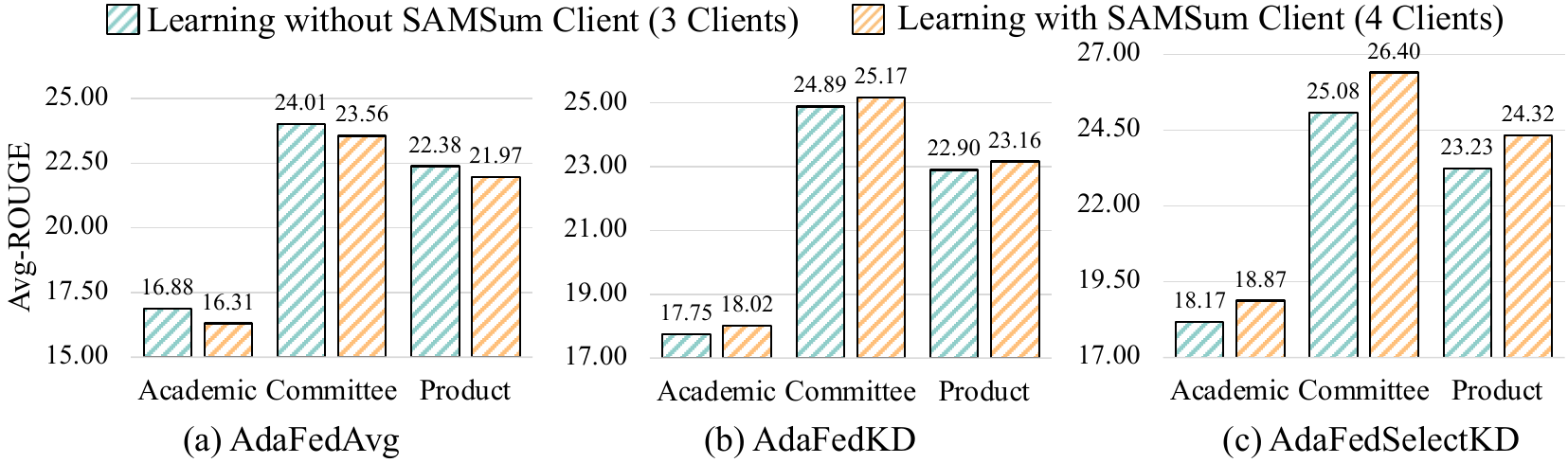}
    \caption{Average ROUGE results based on the extreme non-IID and unbalanced data setting. ``Learning without SAMSum client'' means the original three clients participate in the federated learning process while ``Learning with SAMSum client'' means there are four clients in total with the newly-added SAMSum client joining the learning process.}
    \label{fig:extream}
\end{figure}

To further verify the effectiveness and robustness of our method, we set up an extreme non-IID and unbalanced data distribution setting to assess the performance of different methods. 
To this end, we employ the SAMSum dialogue summarization dataset \cite{samsum} and establish the fourth client, which will participate in the federated learning process along with the previous three clients.
Specifically, SAMSum is a widely-used dataset for the dialogue summarization task, which is vastly different from QMSum. The number of instances (more than 16000 instances), topics (in various scenes of real life), the length of the dialogue (120 tokens on average), the length of the summary (23 tokens on average) and the number of turns (11 turns on average) all differ greatly from QMSum. 
We conduct experiments leveraging \textsc{AdaFedAvg}, \textsc{AdaFedKD} and \textsc{AdaFedSelectKD}.
The results are shown in Figure~\ref{fig:extream}.
Firstly, according to Figure~\ref{fig:extream}(a), we find the previous three clients do not benefit from the newly introduced SAMSum client and actually perform worse.
This is in line with the previous conclusion that the federated averaging algorithm has severe limitations in the presence of non-IID data.
Secondly, on the contrary, knowledge distillation-based federated learning algorithms exhibit their advantages under this challenging setting, with improvements regarding the ROUGE score, as shown in Figure~\ref{fig:extream}(a) and (b).
Thirdly, it is worth noting that \textsc{AdaFedSelectKD} achieves the best results, demonstrating its robustness and efficacy when dealing with extreme non-IID data.

\subsubsection{Client Sampling Setting}

\begin{figure}[t]
    \centering
    \includegraphics[scale=0.41]{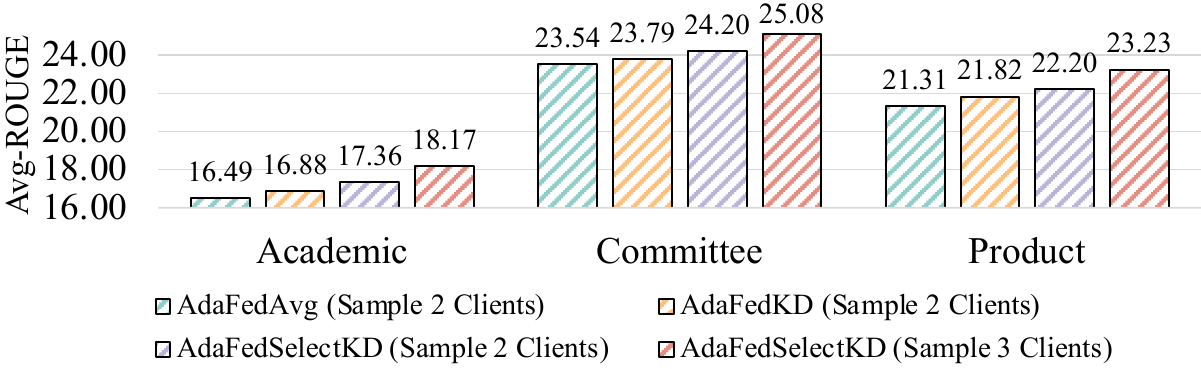}
    \caption{Average ROUGE score based on the client sampling setting.  ``Sample 2 clients'' means that during each round of learning, 2 clients are randomly selected to participate in the learning process.}
    \label{fig:sample}
\end{figure}

In this scenario, we simulate a more pragmatic federated learning environment, in which only a subset of clients participate in each round of the learning process. 
Specifically, we set the participation rate to 70\%, meaning that in our setup, two clients are randomly chosen to participate in the learning procedure during each round.
The results are illustrated in Figure \ref{fig:sample}.
Firstly, the experiment shows that utilizing all three clients --- academic, committee and product --- leads to better performance than learning from only two clients.
Secondly, our proposed method \textsc{AdaFedSelectKD} consistently and stably outperforms the other baseline methods.

\begin{figure}[t]
    \centering
    \includegraphics[scale=0.26]{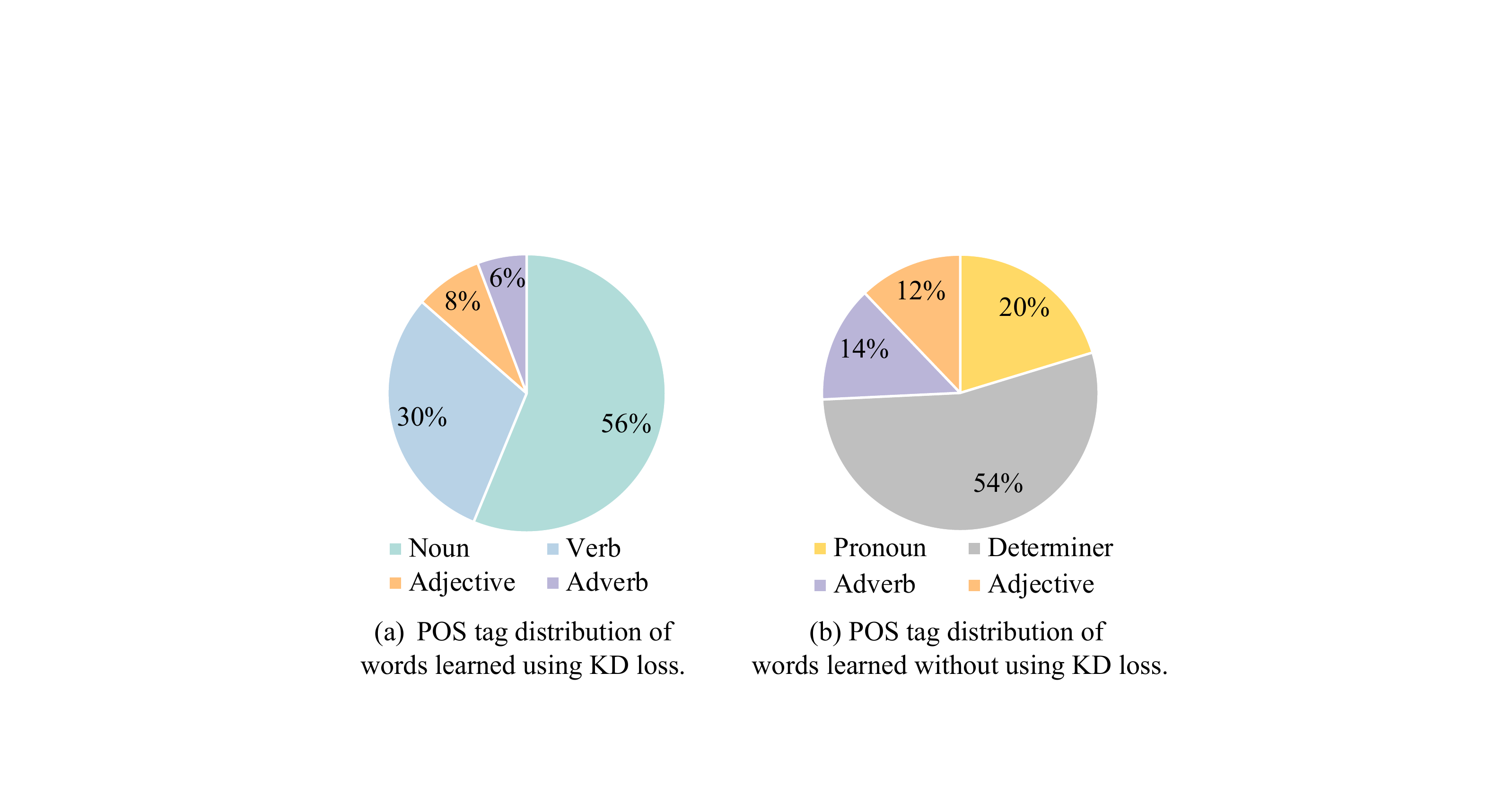}
    \caption{Part-of-speech tag distribution for academic target summary words learned (a) using knowledge distillation loss and (b) without using knowledge distillation loss.}
    \label{fig:pos}
\end{figure}

\subsection{Research Question 3}

To answer the third research question ``\textit{How does the proposed selective knowledge distillation strategy work specifically and what are the underlying mechanisms?}'', we assess the training process by examining whether appropriate target summary words are learned using knowledge distillation and their part-of-speech (POS) tag distribution.

For each client, we extract all target summary words trained from the first round until the final, optimally-performing round to determine whether each summary word adopts knowledge distillation loss and its POS tag information.
For example, at the first client, suppose there is only one training instance with the corresponding summary ``project manager decides to use lcd'', and after two rounds of training, the model achieves its best performance with three target summary words: manager (once), decides (twice).  These are optimized using the knowledge distillation loss.  
Therefore, the proportion of words optimized via knowledge distillation is $3/(2\times6)=25\%$, where 2 is the number of learning rounds and 6 is the number of words in the sentence.

Accordingly, our statistics show that knowledge distillation loss is 86\%, 88\%, and 89\% utilized in generating the target summary words during the training process for the academic, committee, and product clients, respectively.
Furthermore, we calculate the part-of-speech tag distribution of target summary words learned with and without knowledge distillation.
Figure~\ref{fig:pos} illustrates the outcomes for the academic client, while the other two clients exhibit similar distributions. 
The observations suggest that nouns and verbs constitute nearly all of the words optimized through knowledge distillation loss, aligning with the intuition that both nouns and verbs are essential for articulating the core and domain-specific ideas of the meetings.
In contrast, determiners and pronouns make up 74\% of all words learned without using knowledge distillation loss. 

\begin{table*}[t]
\caption{Meeting summaries from our three domains generated by different methods.}
\label{tab:case}
\centering
    \begin{tabular}{ll}
        \toprule 
        \textbf{Method} & \textbf{Meeting Summaries} \\
        \midrule
        \multicolumn{2}{c}{Academic} \\
        \midrule
        \textsc{AdaFedAvg} & \makecell[l]{PhD C explained that there were various delays with different components along the processing chain.}
         \\
        \cdashlinelr{1-2}
        \textsc{AdaFedSelectKD} & \makecell[l]{PhD C said that there were delays of 100ms for silence, 40ms at the input, and 10ms from LDA filters.}   \\
        \cdashlinelr{1-2}
        \textsc{Gold Standard} & \makecell[l]{PhD C explained that the silence probabilities had a 100ms delay, the delta at the input had \\ a 40ms delay, and a 10ms delay was created by LDA filters.} \\
        \midrule
        \multicolumn{2}{c}{Committee} \\
        \midrule
        \textsc{AdaFedAvg} & \makecell[l]{The governing organization developed the framework and simplified the options for \\ resolving issues outside of legal proceedings. } \\
        \cdashlinelr{1-2}
        \textsc{AdaFedSelectKD} & \makecell[l]{The National Police Chiefs’ Council streamlined out-of-court disposals by developing the police approach.}   \\
        \cdashlinelr{1-2}
        \textsc{Gold Standard} & \makecell[l]{The National Police Chiefs' Council was responsible for developing the police approach to out-of-court \\ disposals and simplifying the range of out-of-court disposals.}   \\
        \midrule
        \multicolumn{2}{c}{Product} \\
        \midrule
        \textsc{AdaFedAvg} & \makecell[l]{Industrial Designer proposed an eco-friendly option but Project Manager agreed more with \\  the commercially-appealing proposal.} \\
        \cdashlinelr{1-2}
        \textsc{AdaFedSelectKD} & \makecell[l]{The Industrial Designer suggested solar panel and rechargeable batteries but the Project Manager \\  preferred Project Manager's cradle idea.}  \\ 
        \cdashlinelr{1-2}
        \textsc{Gold Standard} & \makecell[l]{Industrial Designer proposed to incorporate a solar panel and rechargeable batteries, but Project Manager \\  agreed more with Marketing's proposal to include a cradle.}   \\
        \bottomrule
    \end{tabular}
\end{table*}

\begin{figure}[t]
    \centering
    \includegraphics[scale=0.41]{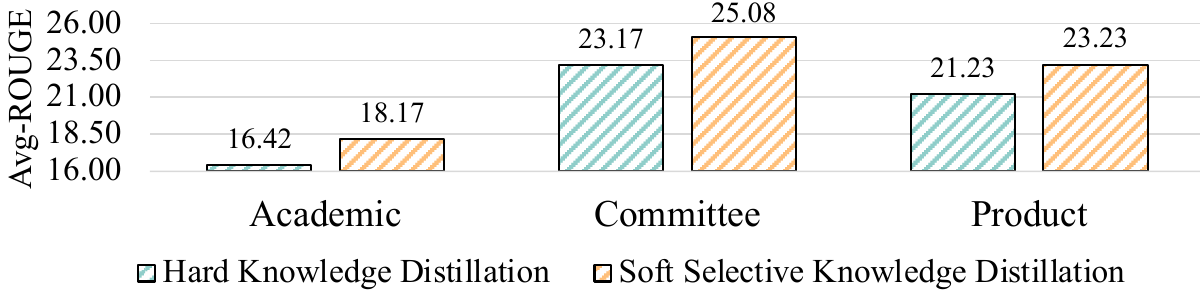}
    \caption{Averaged ROUGE results based on the noun--verb-based hard knowledge distillation and our proposed selective knowledge distillation strategy.}
    \label{fig:soft}
\end{figure}

This distribution insight on the target summary words elicits a natural follow-up: ``\textit{How effective would it be to use knowledge distillation by absolute means only when learning target nouns and verbs?''}. 
To address this question, we conduct experiments by exclusively adopting knowledge distillation rigorously upon learning nouns and verbs.
Figure \ref{fig:soft} shows the results.
We can clearly find that our \textsc{AdaFedSelectKD}, which adaptively makes the distillation decision exhibits superior performance compared to the hard distillation method, demonstrating the necessity of our designed selective strategy.

\subsection{Case Study}

Table \ref{tab:case} illustrates summaries of meetings from three domains generated using various methodologies.
We observe that the baseline method, \textsc{AdaFedAvg}, consistently produces generic meeting summaries lacking in detailed information.
In contrast, our proposed method, \textsc{AdaFedSelectKD}, yields summaries that are more informative and tailored to the domain. 
Moreover, the gold standard meeting summaries continue to demonstrate advantages in conciseness and informativeness, highlighting the challenges intrinsic to meeting summarization.

\section{Related Work}

\subsection{Meeting Summarization}
Meeting summarization \cite{Feng2021ASO,Kumar2022MeetingSA,Rennard2022AbstractiveMS} aims to pack crucial information of a given meeting into a concise yet comprehensive summary highlighting the most salient points.
In addition to the challenges inherent in traditional summarization tasks, meeting summarization must address unique difficulties arising from its multi-participant nature.
To facilitate progress in this domain, various datasets have been curated \cite{ami,icsi,Zhong2021QMSumAN,Hu2023MeetingBankAB,wu2023vcsum}, enabling the development of state-of-the-art models that incorporate versatile knowledge \cite{Feng2020DialogueDG,Zhu2020AHN,Riedhammer2008AKB,Zhong2021DialogLMPM,Zhang2021AnES,Liu2022DynamicSW,zhang-etal-2022-summn} and achieve the best results.
However, privacy concerns, which are inextricably intertwined with meeting content, have received little attention in the literature, hampering real-world application. 
Lee and Sogaard \cite{Lee2023PrivateMS} take the initiative to address this issue by exploring differential privacy (DP) \cite{Dwork2014TheAF} for meeting summarization, focusing primarily on a single domain.
In this work, we conduct the first systematic study of meeting summarization under the federated learning framework, accounting for the heterogeneity and unbalance of data across multiple domains.

\subsection{Federated Learning}
Federated learning enables collaborative machine learning without the centralized collection of potentially sensitive raw data, thereby paving the way for stronger privacy guarantees when building predictive models \cite{McMahan2016CommunicationEfficientLO}.
With mounting concerns regarding privacy issues, this paradigm has garnered significant research interest including diverse research directions \cite{Yang2019FederatedML}.
In particular, owing to the inevitable inclusion of private information in texts, a variety of studies have explored diverse natural language processing tasks within the federated learning framework \cite{Liu2021FederatedLM}.
The predominant efforts in this realm have focused on natural language understanding tasks, such as spoken language understanding \cite{Huang2020FederatedLF} and text classification \cite{Li2022FederatedSB}.
Recent years have witnessed a trend toward applying the federated learning framework to natural language generation tasks \cite{Lin2021FedNLPBF}.
Our work follows this line of work and is the first to explore the federated multi-domain meeting summarization task.

\subsection{Parameter-efficient Fine-tuning}
The field of natural language processing is currently dominated by large language models  \cite{gpt3}. 
Despite their superiority, fine-tuning all the parameters of these immense models on various downstream tasks becomes prohibitively complicated as both model size and number of tasks increase \cite{He2021TowardsAU}.
To alleviate this problem, parameter-efficient fine-tuning is coming to the rescue by updating only a small number of extra parameters while keeping most pre-trained parameters frozen \cite{Houlsby2019ParameterEfficientTL,Li2021PrefixTuningOC}.
Opportunely, such lightweight alternatives are well-suited for reducing communication costs in the federated learning framework. 
Based on this foundation, the amalgamation of federated learning and parameter-efficient fine-tuning unveils vast potential for diverse applications \cite{Zhao2022FedPromptCA}.
In this paper, we craft two types of adapter modules, a global adapter and a local adapter, which collaboratively and efficiently facilitate federated client--server communication.

\subsection{Knowledge Distillation}
Knowledge distillation refers to the process of transferring knowledge from a teacher model to a student model without significant performance degradation. It has proven to be an effective method for improving model performance \cite{Hinton2015DistillingTK}. 
In recent years, knowledge distillation has been applied in the federated learning framework and has demonstrated its ability to mitigate the effects of data heterogeneity \cite{Mora2022KnowledgeDF}.
Our proposed framework builds upon this and takes one step further to explore the combination of both knowledge distillation and parameter-efficient fine-tuning while introducing one carefully designed selective strategy to enable an adaptive learning process.

\section{Limitations and Potential Advancements}
The aforementioned experiments are conducted using the dataset provided by the research community and based on a simulated federated environment.   This may not fully reflect the complexities of real-world scenarios and could potentially lead to two limitations: \\

{\bf 1. Well-curated dataset exhibits less variability.}
In reality, meetings, even those within the same domain, vary with respect to their participants, discussion topics, and duration. Moreover, meeting transcripts are typically generated via automatic speech recognition (ASR) systems, resulting in noisy, imperfect textual data. Useful forms of meeting summaries also depend on the target user needs, spanning full-text summaries, highlighted extracts, identification of action items, and more. While our research attempts to address real-world limitations, the dataset employed in our experiment likely does not adequately capture the complexities of real-world meeting settings.
We envision that future advancements can develop more appropriate benchmarks, design more comprehensive experimental paradigms, and possibly even collaborate with corporations to narrow the divide between research explorations and practical real-world applications. 

{\bf 2. Simulated federated environment lacks uncertainty.}
In practical federated learning deployments, addressing the challenges of non-IID and unbalanced data is still insufficient. It is also imperative to overcome various issues stemming from communication uncertainties, such as client--server latency, asynchronous client learning updates, and client dropout. These factors are unaddressed in the current research, and present challenges to any federated learning techniques. 
As such, future advancements could conduct larger-scale decentralized experiments in which the federated learning procedure is intentionally perturbed to stress-test such a framework's robustness to uncertainty, thereby approximating real-world conditions more closely. 

\section{Conclusion and Future Work}

We examine the multi-domain meeting summarization task under a federated learning paradigm.  We show that this represents a more pragmatic and realistic configuration than prior work on learning meeting summary models over centralized meeting data.
Moreover, to mitigate two challenges, namely limited server--client communication and the non-IID data learning situation, we propose a unified method, \textsc{AdaFedSelectKD}, which succeeds in reducing communication costs and addressing the domain drift problem of the client model. 
Through comprehensive empirical studies, our method demonstrates its effectiveness and robustness that can achieve comparable results with centralized training methods while exhibiting its superiority in handling the intricacies of non-IID data.

We believe that future work will strive to apply the proposed method to real scenarios. 
In our own work, we aim to craft data resources and design experimental settings to adequately simulate real-world federated learning circumstances.
We plan to collaborate with organizations to implement the federated learning framework and evaluate our proposed method in addressing a variety of exigencies.
Further investigation will build from this foundation, to incorporate differential privacy techniques to further augment our model's privacy preservation characteristics.
These advances promise to make \textsc{AdaFedSelectKD} a practical solution for meeting summarization. 

\section*{Acknowledgments}
This work was supported by National Key R\&D Program of China via grant 2020AAA0106502, National Natural Science Foundation of China (NSFC) via grant 62276078, Key R\&D Program of Heilongjiang via grant 2022ZX01A32 and the International Cooperation Project of PCL, PCL2022D01.

\bibliographystyle{IEEEtran}
\bibliography{ref}

\vfill

\end{document}